# Deep Interactive Object Selection

Ning Xu
University of Illinois at Urbana-Champaign
ningxu2@illinois.edu

Brian Price
Adobe Research
bprice@adobe.com

Scott Cohen
Adobe Research
scohen@adobe.com

Jimei Yang
Adobe Research
jimyang@adobe.com

Thomas Huang
University of Illinois at Urbana-Champaign
t-huang1@illinois.edu

## Abstract

*Interactive object selection is a very important research problem and has many applications. Previous algorithms require substantial user interactions to estimate the foreground and background distributions. In this paper, we present a novel deep-learning-based algorithm which has a much better understanding of objectness and thus can reduce user interactions to just a few clicks. Our algorithm transforms user-provided positive and negative clicks into two Euclidean distance maps which are then concatenated with the RGB channels of images to compose (image, user interactions) pairs. We generate many of such pairs by combining several random sampling strategies to model users' click patterns and use them to finetune deep Fully Convolutional Networks (FCNs). Finally the output probability maps of our FCN-8s model is integrated with graph cut optimization to refine the boundary segments. Our model is trained on the PASCAL segmentation dataset and evaluated on other datasets with different object classes. Experimental results on both seen and unseen objects clearly demonstrate that our algorithm has a good generalization ability and is superior to all existing interactive object selection approaches.*

## 1. Introduction

Interactive object selection (also known as interactive segmentation) has become a very popular research area over the past years. It enables users to select objects of interest accurately by interactively providing inputs such as strokes and bounding boxes. The selected results are useful for various applications such as localized editing and image/video composition.

There are many algorithms proposed to solve this problem. One of the most famous algorithms is proposed by Boykov and Jolly [2] where they formulate interactive segmentation as the graph cut optimization and solve it via max-flow/min-cut energy minimization. Rother *et al.* [19] extend graph cut by using a more powerful, iterative version of optimization. Bai and Sapiro [1] present a new algorithm that computes weighted geodesic distances to the user-provided scribbles. Grady [8] uses the graph theory to estimate the probabilities of random walks from unlabeled pixels to labeled pixels. In order to get accurate segmentation, all these algorithms require substantial user interactions to have a good estimation of the foreground/background distributions. In contrast, our approach simplifies user interactions to a few clicks, with one or two clicks usually giving reasonably good results. The advantage of our approach over the others is the capability to understand objectness and semantics by leveraging deep learning techniques. To our best knowledge, this is the first work that solves interactive segmentation in the framework of deep learning.

Our approach is inspired by recent successes of deep fully convolutional neural networks (FCNs) on the semantic segmentation problem [15, 26, 3, 14, 12]. Long *et al.* [15] adapt popular deep classification networks into FCNs for semantic segmentation and improve the architecture with multi-resolution layer combinations. Built upon this, Chen *et al.* [3] combine the outputs of FCNs with Conditional Random Field (CRF) while Zheng *et al.* [26] formulate mean-field approximate inference as Recurrent Neural Network (RNN) and plug it on top of FCNs to get finer results.

A seemingly plausible transformation of those approaches to interactive segmentation is that we first perform semantic segmentation on the whole image and then select the connected components which contain user-provided selections. However, there exists at least three problems with this approach. First, it is not always clear how to response to use inputs. For example, if the user places a foreground click and background click inside the same class label, this approach cannot response to that. Second, current semantic segmentation methods do not support instance-level seg-mentation while that is often the user's desire. Last but not the least, current semantic segmentation approaches do not generalize to unseen objects. This means that we have to train a model for every possible object in the world, which is obviously impractical.

In this paper, we present a novel algorithm for interactive object selection (Fig. 1). To select an object in an image, users provide positive and negative clicks which are then transformed into separate Euclidean distance maps and concatenated with the RGB channels of the image to compose a (image, user interactions) pair. FCN models are fine tuned on many of these pairs generated by random sampling. Moreover, graph cut optimization is combined with the outputs of our FCN models to get satisfactory boundary localization. The key contributions of this paper are summarized as follows:

- We propose an effective transformation to incorporate user interaction with current deep learning techniques.

- We propose several sampling strategies which can represent users' click behaviors well and obtain the required training data inexpensively.

- Our interactive segmentation system is real time given a high-end graphics processing units (GPU).

The rest of the paper is organized as follows. Section 2 gives a brief review of related works. The proposed algorithm is elaborated in Section 3. Experimental results are presented in Section 4 and finally we conclude the paper in Section 5.

## 2. Related works

Interactive segmentation has been studied for many years. There are many interactive approaches, such as contour-based methods [17, 10] and bounding box methods [19]. Stroke-based methods are popular, and use a number of underlying algorithms, including normalized cuts [20], graph cut [2, 11, 23], geodesics [1, 4], the combination of graph cut and geodesics [18, 9] and random walks [8]. However, all these previous algorithms estimate the foreground/background distributions from low-level features. Unfortunately, low-level features are insufficient at distinguishing the foreground and background in many cases, such as in images with similar foreground and background appearances, complex textures and appearances, and difficult lighting conditions. In such cases, these methods struggle and require excessive user interaction to achieve desirable results. In contrast, our FCN model is trained end-to-end and has a high level understanding of objectness and semantics, therefore simplifying user interactions to just a few clicks.

The task of semantic segmentation is closely related to interactive segmentation. Many algorithms have been proposed in the past [25, 21, 22]. Due to the great improvements on image classification and detection by deep neural networks especially the convolutional neural networks (CNNs), many researchers have recently applied CNNs to the problem of semantic segmentation. Farabet *et al.* [6] use a multi-scale convolutional network trained from raw pixels for scene labeling. Girshick *et al.* [7] apply CNNs to bottom-up regions proposals for object detection and segmentation and improve over previous low-level-feature-based approaches greatly. Long *et al.* [15] adapt high-capacity CNNs to FCNs which can be trained end-to-end, pixels-to-pixels and leverage a skip architecture which combines model responses at multiple layers to get finer results. However, as explained in the introduction, semantic segmentation is not directable for interactive segmentation. Our model is based on FCNs but different from [15] in mainly two points. 1) Our model is trained on randomly generated (image, user interactions) pairs which are the concatenations of RGB channels and transformed Euclidean distance maps. 2) Our model has only two labels – "object" and "background".

Other work has looked at improving the boundary localization of CNN semantic segmentation approaches. Chen *et al.* [3] combine the outputs of FCNs with fully connected CRF. Zheng *et al.* [26] formulate mean-field approximate inference as RNNs and train with FCNs end-to-end. They improve the mean intersection over union (IU) accuracy of FCNs from 62.2% to 71.6% and 72% respectively. Although our FCN models are quite general to be combined with their approaches, their segmentation results are far less acceptable for the interactive segmentation task. Therefore, we propose a simple yet effective approach that combine graph cut optimization with our FCN output maps, which enables our algorithm achieve high IU accuracy with even a single click.

## 3. The proposed algorithm

We propose a deep-learning-based algorithm for interactive segmentation. User interactions are first transformed into Euclidean distance maps and then concatenated with images' RGB channels to fine tune FCN models. After the models are trained, graph cut optimization is combined with the probability maps of FCN-8s to get the final segmentation results. Figure 1 illustrates the framework of how we train our FCN models.

### 3.1. Transforming user interactions

In our approach, a user can provide positive and negative clicks (or strokes) sequentially in order to segment objects of interest. A click labels a particular location as being either "object" or "background". A sequence of user inter-actions $\mathcal{S}$ includes a positive click set $\mathcal{S}^1$ which contains all user-provided positive clicks and a negative click set $\mathcal{S}^0$ which contains all user-provided negative clicks. Our algorithm uses a Euclidean distance transformation to transform $\mathcal{S}^1$ and $\mathcal{S}^0$ to separate channels $\mathbf{U^1}$ and $\mathbf{U^0}$ respectively. Each channel is a 2D matrix with the same height and width as the original image. To calculate the pixel value $u_{ij}^t$ at the location $(i,j)$, $t \in \{0,1\}$, let us first define an operator $f$ such that given a set of points $p_{ij} \in \mathcal{A}$ where $(i,j)$ is the point location, then for any point $p_{mn}$, $f(p_{mn}|\mathcal{A}) = \min_{\forall p_{ij} \in \mathcal{A}} \sqrt{(m-i)^2 + (n-j)^2}$. In other words, the operator $f$ calculates the minimum Euclidean distance between a point and a set of points. Then,

$$u_{ij}^t = f(p_{ij}|\mathcal{S}^t), \quad t \in \{0,1\} \quad (1)$$

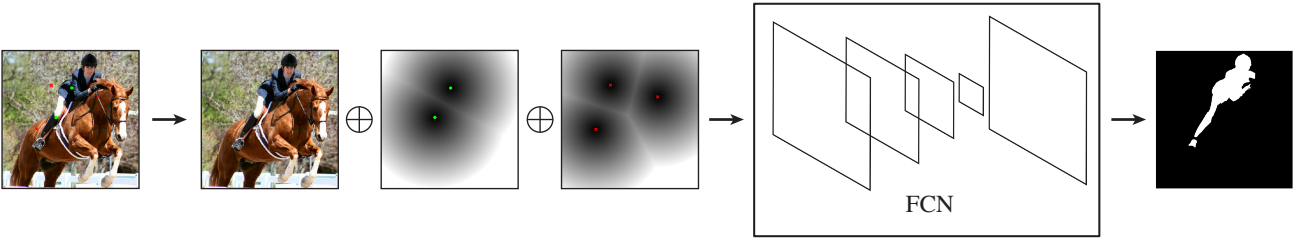

Figure 1: The framework of learning our FCN models. Given an input image and user interactions, our algorithm first transforms positive and negative clicks (denoted as green dots and red crosses respectively) into two separate channels, which are then concatenated (denoted as ⊕) with the image's RGB channels to compose an input pair to the FCN models. The corresponding output is the ground truth mask of the selected object.

For the efficiency of data storage, we truncate $u_{ij}^t$ to 255. It should be noted that it is possible that $\mathcal{S}^0$ is a empty set since in many scenarios our algorithm has perfect segmentation results with even one single positive click. In this case, all $u_{ij}^0$ are set to 255. Then we concatenate the RGB channels of the image with $\mathbf{U^1}, \mathbf{U^0}$ to compose a (image, user interaction) pair.

### 3.2. Simulating user interactions

It should be noted that different users tend to have different interaction sequences for selecting the same object. Therefore our FCN models need a lot of such training pairs to learn this. However, it is too expensive to collect many interaction sequences from real users. We thus use random sampling to automatically generate those pairs. Let $\mathcal{O}$ be the set of ground truth pixels of the object and let us define a new set $\mathcal{G} = \{p_{ij}|p_{ij} \in \mathcal{O} \text{ or } f(p_{ij}|\mathcal{O}) \geq d\}$. Let $\mathcal{G}^c$ denote the complementary set of $\mathcal{G}$. It is easy to see that the pixels in $\mathcal{G}^c$ have two properties: 1) they are background pixels and 2) they are within a certain distance range to the object. To sample positive clicks, we randomly select $n$ pixels in $\mathcal{O}$ where $n \in [1, N_{pos}]$. The pixels in $\mathcal{O}$ are actually filtered in the way that 1) any two pixels are at least $d_{step}$ pixels away from each other and 2) any pixel is at least $d_{margin}$ pixels away from the object boundaries.

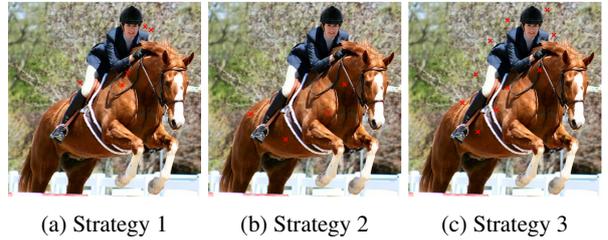

(a) Strategy 1    (b) Strategy 2    (c) Strategy 3

Figure 2: A visual example of the three sampling strategies for negative clicks. The person is the foreground object.

To sample negative clicks, we combine several sampling strategies to model the complexity of users' click patterns.

- **Strategy 1:** $n$ negative clicks are randomly sampled in the set $\mathcal{G}^c$, where $n \in [0, N_{neg1}]$. $\mathcal{G}^c$ is filtered in the same way as $\mathcal{O}$.

- **Strategy 2:** $n_i$ negative clicks are randomly sampled on each negative object $\mathcal{O}_i$ in the same image, where $n_i \in [0, N_{neg2}]$. Each $\mathcal{O}_i$ is filtered in the same way as $\mathcal{O}$.

- **Strategy 3:** $N_{neg3}$ negative clicks are sampled to cover the outside object boundaries as much as possible. In detail, the first negative click is randomly sampled in $\mathcal{G}^c$. Then the following clicks are obtained sequentially by

$$p_{next} = \arg \max_{p_{ij} \in \mathcal{G}^c} f(p_{ij}|\mathcal{S}^0 \cup \mathcal{G}) \quad (2)$$

where $\mathcal{S}^0$ includes all previously sampled negative clicks.

Figure 2 presents an example of the three strategies. The sampled negative clicks from **Strategy 1** or **2** alone do not always follow users' typical click patterns, therefore making them harder for our models to learn. The sampled negative clicks from **Strategy 3** surround the object evenly, which has a strong pattern but is easy to learn. We find that using all three stategies provides better results than relying on any one strategy, therefore we combine them together. Specifically, for each object in an image we randomly sample $N_{\text{pairs}}$ training pairs of (image, user interactions). Each pair is generated by one of the sampling strategies with an equal probability.

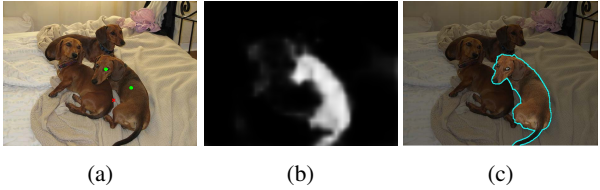

(a)　　　　　　(b)　　　　　　(c)

Figure 3: An example of Section 3.4. (a) An testing image and user interactions. (b) The output probability map from FCN-8s. (c) The result after graph cut.

### 3.3. Fine tuning FCN models

We leverage FCNs to learn the interactive segmentation task. The training samples to our models are (image, user interactions) pairs and the labels are the binary masks of corresponding objects. We first fine tune a stride-32 FCN model (FCN-32s) from the stride-32 semantic segmentation model of [15]. For the two extra channels of filters in the first convolutional layer, we use zero initialization. We also tried initialization with the mean value of those filter weights, but it shows no difference. After fine tuning FCN-32s, we continue to fine tune a stride-16 FCN (FCN-16s) from FCN-32s with the same training data. Finally we fine tune a stride-8 FCN (FCN-8s) model from FCN-16s. As suggested by [15], training finer-stride FCNs does not provide further benefits, which we also observed.

It takes approximately three days to fine tune FCN-32s and five days to fine tune FCN-16s and FCN-8s. By balancing the trade-offs between the performance and time, each FCN model is trained about 20 epochs. FCN-32s converges fast in the first two epochs while a longer training time gives finer segmentation results. We also find that FCN-16s has obvious improvements over FCN-32s especially in regions close to object boundaries, but the accuracy of FCN-16s and FCN-8s are similar.

### 3.4. Graph cut optimization

From the outputs at the last layer of FCN-8s we can obtain a probability map $\mathbf{Q}$, of which the entry $q_{ij}$ indicates how likely the pixel $p_{ij}$ is labeled as "object" (*e.g.* Figure 3b). Directly thresholding $q_{ij}$ at 0.5 gives us very coarse segmentation masks, which are not useful for interactive segmentation. Instead, we integrate $\mathbf{Q}$ into the graph cut optimization [2]:

$$E(L) = \lambda \cdot R(L) + B(L) \quad (3)$$

Where $\lambda$ is a coefficient that specifies a relative importance between $R(L)$ and $B(L)$.

The first term $R(L) = \sum_{p_{ij} \in \mathcal{P}} R_{p_{ij}}(L_{p_{ij}})$, where $R_{p_{ij}}(L_{p_{ij}})$ estimates the penalty of assigning pixel $p_{ij}$ to label $L_{p_{ij}}$. Our algorithm defines

$$R_{p_{ij}}(L_{p_{ij}}) = \begin{cases} -\log(q_{ij}), & \text{if } L_{p_{ij}} = \text{"object"} \\ -\log(1 - q_{ij}), & \text{otherwise} \end{cases} \quad (4)$$

The second term $B(L) = \sum_{\{p_{ij}, p_{mn}\} \in \mathcal{N}} B_{\{p_{ij}, p_{mn}\}} \cdot \delta(L_{p_{ij}}, L_{p_{mn}})$, where $B_{\{p_{ij}, p_{mn}\}}$ comprises the properties of object boundaries. Our algorithm defines

$$B_{\{p_{ij}, p_{mn}\}} \propto exp(-\frac{(I_{p_{ij}} - I_{p_{mn}})^2}{2\sigma^2}) \cdot \frac{1}{dist(p_{ij}, p_{mn})} \quad (5)$$

Our algorithm solves Equation 3 via max-flow/min-cut energy minimization. Figure 3c illustrates the result after graph cut optimization.

### 3.5. Evaluation and complexity

A user can provide positive and negative clicks sequentially to select objects of interest. Each time a new click is added, our algorithm recomputes the two distance maps $\mathbf{U^1}$ and $\mathbf{U^0}$. Then the new (image, user interactions) pair is sent to our FCN-8s model and a new probability map $\mathbf{Q}$ is obtained. Graph cut uses $\mathbf{Q}$ to update the segmentation results without recomputing everything from scratch. To compare our algorithm with other approaches, we also design a method to automatically add a click given the current segmentation mask and the ground truth mask. The method places a seed at the mislabeled pixel that is farthest from the boundary of the current selection and the image boundaries, mimicing a user's behavior under the assumption that the user clicks in the middle of the region of greatest error.

Given high-end GPUs like NVIDIA Titan X, the computation of $\mathbf{Q}$ is very fast and less than 100 millisecond. Graph cut optimization is also very efficient on modern CPUs. Therefore our algorithm satisfies the speed requirement for the interactive segmentation task.

## 4. Experiments

### 4.1. Settings

We fine tune our FCN models on the PASCAL VOC 2012 segmentation dataset [5] which has 20 distinct object categories. We use its 1464 training images which have instance-level segmentation masks and their flipped versions to sample the (image, user interactions) pairs. The choices of some sampling hyper-parameters are: $d$ is set to be 40, $N_{\text{pos}}$ is set to be 5, $N_{\text{neg1}}, N_{\text{neg2}}, N_{\text{neg3}}$ are set to be 10, 5 and 10 respectively. $N_{\text{pairs}}$ is set to be 15. The total number of sampled training pairs is about 80k. 200 validation images are randomly sampled from the whole training set to control the learning of our models.

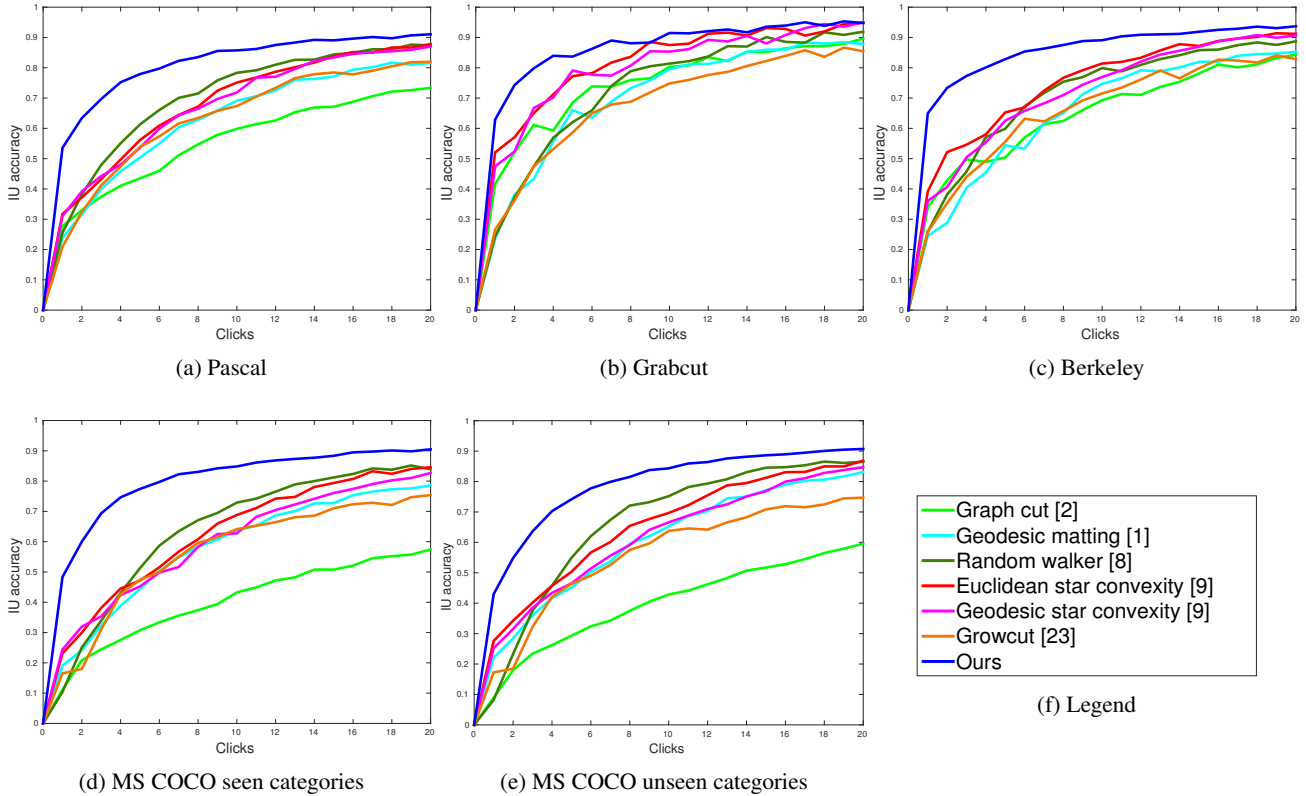

Figure 4: The mean IU accuracy *vs.* the number of clicks on the (a) Pascal (b) Grabcut (c) Berkeley (d) MS COCO seen categories and (e) MS COCO unseen categories datasets. The legend of these plots is shown in (f).

We compare our algorithm to several popular interactive segmentation algorithms [2, 1, 8, 9, 23]. Since the other algorithms cannot estimate foreground/background distributions with a single click, we enlarge every click to a big dot with a radius 5 for them. We use such big dots for our graph cut refinement but only use single clicks for our FCN models. To evaluate, we record the updated IU accuracy of an object given sequential clicks which are automatically generated in the way described in Section 3.5. The maximum number of clicks on a single object is limited to 20. We also record how many clicks are required to achieve a certain IU accuracy for the object. If the IU accuracy cannot be achieved in 20 clicks, we will threshold it by 20. Finally, we average each metric over all objects in a dataset.

### 4.2. Results

We evaluate all the algorithms on four public datasets: Pascal VOC 2012 segmentation validation set, Grabcut [19], Berkeley [16] and MS COCO [13]. The quantitative results of the two metrics on different datasets are shown in Figure 4 and Table 1 respectively.

**Pascal:** The validation set has 1449 images and many of them contain multiple objects. From Figure 4a we can see that our algorithm is better than all the other algorithms. Since the validation set contains 20 object categories which have been seen in our training set, we test our algorithm on other datasets with different objects to prove the generalization capability of our algorithm to unseen object classes.

**Grabcut** and **Berkeley:** These two datasets are benchmark datasets for interactive segmentation algorithms. On the Grabcut dataset (Figure 4b), our algorithm achieves better results with a few clicks and has a similar IU accuracy with Geodesic/Euclidean star convexity [9] with more clicks. Since Grabcut only has 50 images and most images have distinct foreground and background distributions which can be handled well by low-level-feature-based algorithms, our advantage over other methods is smaller than it is on more challenging datasets. On the Berkeley dataset (Figure 4c), our algorithm achieves better IU accuracy at every step and increases the IU accuracy much faster than the others at the beginning of the interactive selection.

| Segmentation models | Pascal (85% IU) | Grabcut (90% IU) | Berkeley (90% IU) | MS COCO seen categories (85% IU) | MS COCO unseen categories (85% IU) |
|---|---|---|---|---|---|
| Graph cut [2] | 15.06 | 11.10 | 14.33 | 18.67 | 17.80 |
| Geodesic matting [1] | 14.75 | 12.44 | 15.96 | 17.32 | 14.86 |
| Random walker [8] | 11.37 | 12.30 | 14.02 | 13.91 | 11.53 |
| Euclidean start convexity [9] | 11.79 | 8.52 | 12.11 | 13.90 | 11.63 |
| Geodesic start convexity [9] | 11.73 | 8.38 | 12.57 | 14.37 | 12.45 |
| Growcut [23] | 14.56 | 16.74 | 18.25 | 17.40 | 17.34 |
| Ours | **6.88** | **6.04** | **8.65** | **8.31** | **7.82** |

Table 1: The mean number of clicks required to achieve a certain IU accuracy on different datasets by various algorithms. The IU accuracy for different datasets is indicated in the parentheses. The best results are emphasized in **bold**.

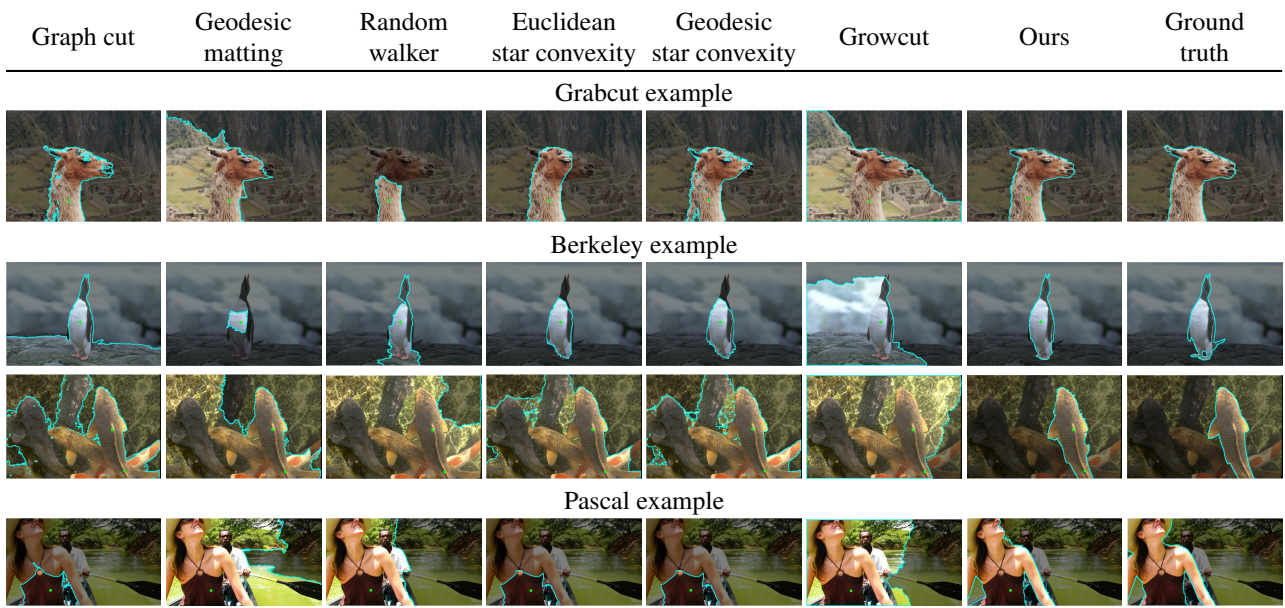

Figure 5: The segmentation results by different algorithms given the same user interaction sequences. Each row is an testing image from one dataset. Each of the first seven columns represent the segmentation results by one algorithm and the rightmost column shows the ground truths. In each figure, green dots indicate positive clicks. Background regions are faded to black and object boundaries are outlined in cyan.

**MS COCO:** MS COCO is a large-scale segmentation dataset and has 80 object categories. 60 of them are distinct from the Pascal dataset. We randomly sample 10 images per categories and test all the algorithms on the 20 seen categories and 60 unseen categories separately. Our algorithm still consistently performs better than the other algorithms by a large margin in both cases.

Our algorithm also requires the least number of clicks to achieve a certain IU accuracy on all the datasets. Figure 4 and Table 1 clearly demonstrate that 1) our algorithm achieves more accurate results with less interaction than other methods and 2) our algorithm has a good generalization ability to all kinds of objects. Given the same user interaction sequences, some segmentation results by different algorithms are illustrated in Figure 5 and 9. In many examples, our algorithm obtains very good results in just one single click while the others either only segment a part of the object or completely fail. This is because our FCN models have a high-level understanding of the objectness and semantics, in contrary to the other approaches simply relying on low-level features. We also show a failed segmentation result by our algorithm in Figure 9. The failure is because FCNs cannot capture thin structures and fine details very well. Therefore the output probabilities from our FCN-8s model are not accurate enough in those areas, which affects the performance of graph cut in producing our final result.

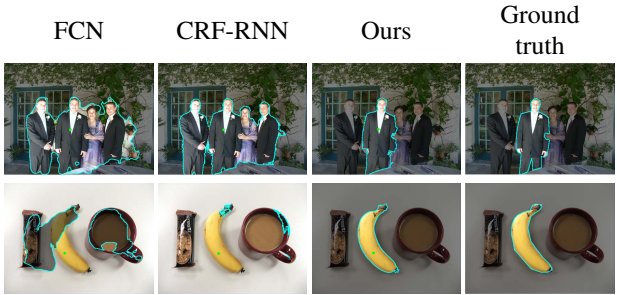

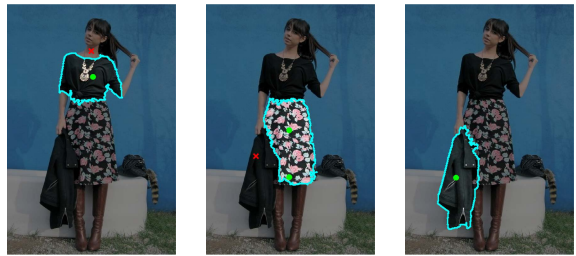

Figure 6: The segmentation results by the semantic segmentation algorithms (FCN and CRF-RNN) and our algorithm. The first row is a testing image from seen categories (*i.e.* "person"). The second row is a testing image from unseen categories (*i.e.* "banana").

Figure 7: The segmentation results of clothing parts by our algorithm on the Fashionista dataset. The clothing parts from the left image to the right image are "shirt", "skirt" and "jacket".

| Segmentation models | MS COCO seen categories | MS COCO unseen categories |
| --- | --- | --- |
| FCN [15] | 42.37% | 16.14% |
| CRF RNN [26] | 47.01% | 13.28% |
| Ours | **48.35%** | **42.94%** |

Table 2: The mean IU accuracy with a single positive click on the MS COCO seen and unseen categories. The best results are emphasized in **bold**.

### 4.3. Comparisons to semantic segmentation approaches

Since all existing interactive segmentation algorithms are only based on low-level features, we should also compare our algorithm to some models that understand high-level semantics, such as FCNs [15] and CRF-RNN [26]. However, they neither support instance-level segmentation nor can respond to users' interactions. To make them comparable, we design a simple strategy such that the connected component of a given label that contains the user click is selected as foreground and the other areas are treated as background. It is not straightforward how to respond to negative clicks, therefore we only compare results by a single positive click.

The visual comparison results are shown in Figure 6. In the first example, since "person" is a known category to FCN and CRF-RNN, they are able to segment all the persons in the image. But they cannot segment the man in the middle who overlaps with other persons. In the second example, "banana" is a new category to FCN and CRF-RNN. Therefore they don't recognize it at all. Table 2 presents the mean IU accuracy with a single positive click on the MS COCO dataset, which demonstrates the limitations of semantic segmentation approaches directly applied to interactive segmentation. For seen categories, since many of the class instances are non-overlapping, we only have a modest improvement. However, remember that our algorithm was given only one click, and with more clicks we can greatly improve our results. For unseen classes, our algorithm performs significantly better, proving both our ability to generalize to new classes and the effectiveness of our algorithm in combining user interactions with deep learning techniques.

### 4.4. Segmenting object parts

Previous results demonstrate that our algorithm performs very well on general objects. Moreover, although our FCN models are only trained on whole objects, our algorithm can still select their subparts. In Figure 7 we show some segmentation results of clothing parts on the Fashionista dataset [24]. This demonstrates the flexibility of our algorithm and the effectiveness of our learning framework that enables our models to understand users' intentions well. In addition, compared with the other interactive segmentation approaches, there is no doubt that they need many user interactions to achieve the results. Compared with automatic semantic segmentation methods like FCNs, they are trained to segment entire people and thus cannot get the subparts. This again shows the advantages of our algorithm.

### 4.5. Refinement by Graph Cut

We illustrate the differences of segmentation results before and after our graph cut refinement in Figure 8. The first row shows the output probability maps of our FCN-8s model thresholded at 0.5. We can see our model responds correctly to the user interactions and selects most parts of the bus. But the results along object boundaries are not very accurate. Therefore our algorithm leverages graph cut to refine the results. The second row shows the final results of our algorithm. Clearly the results are more satisfactory and have better boundary localization.

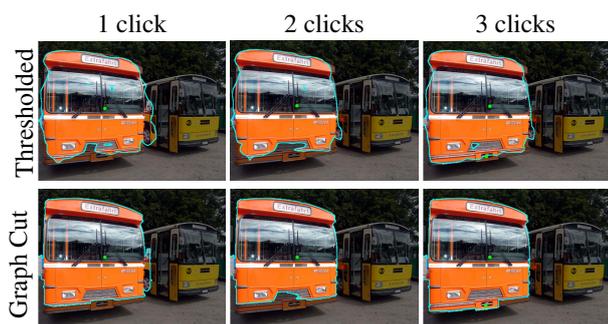

Figure 8: The sequential segmentation results before and after our graph cut refinement. The first row shows the results by thresholding the output probability maps of our FCN-8s model without using graph cut. The second row shows our final results after graph cut.

## 5. Conclusion

The proposed algorithm is the first work that solves the interactive segmentation problem by combining user interactions with current deep learning models. Our algorithm transforms user interactions to Euclidean distance maps and trains FCN models to recognize "object" and "background" based on many synthesized training samples. Our algorithm also combines graph cut optimization with the output of the FCN-8s model to refine the segmentation results along object boundaries. Experimental results clearly demonstrate the superiority of the proposed deep algorithm over existing interactive methods using hand designed, low level features. Our method can achieve high quality segmentations with a small amount of user effort, often just a few clicks.

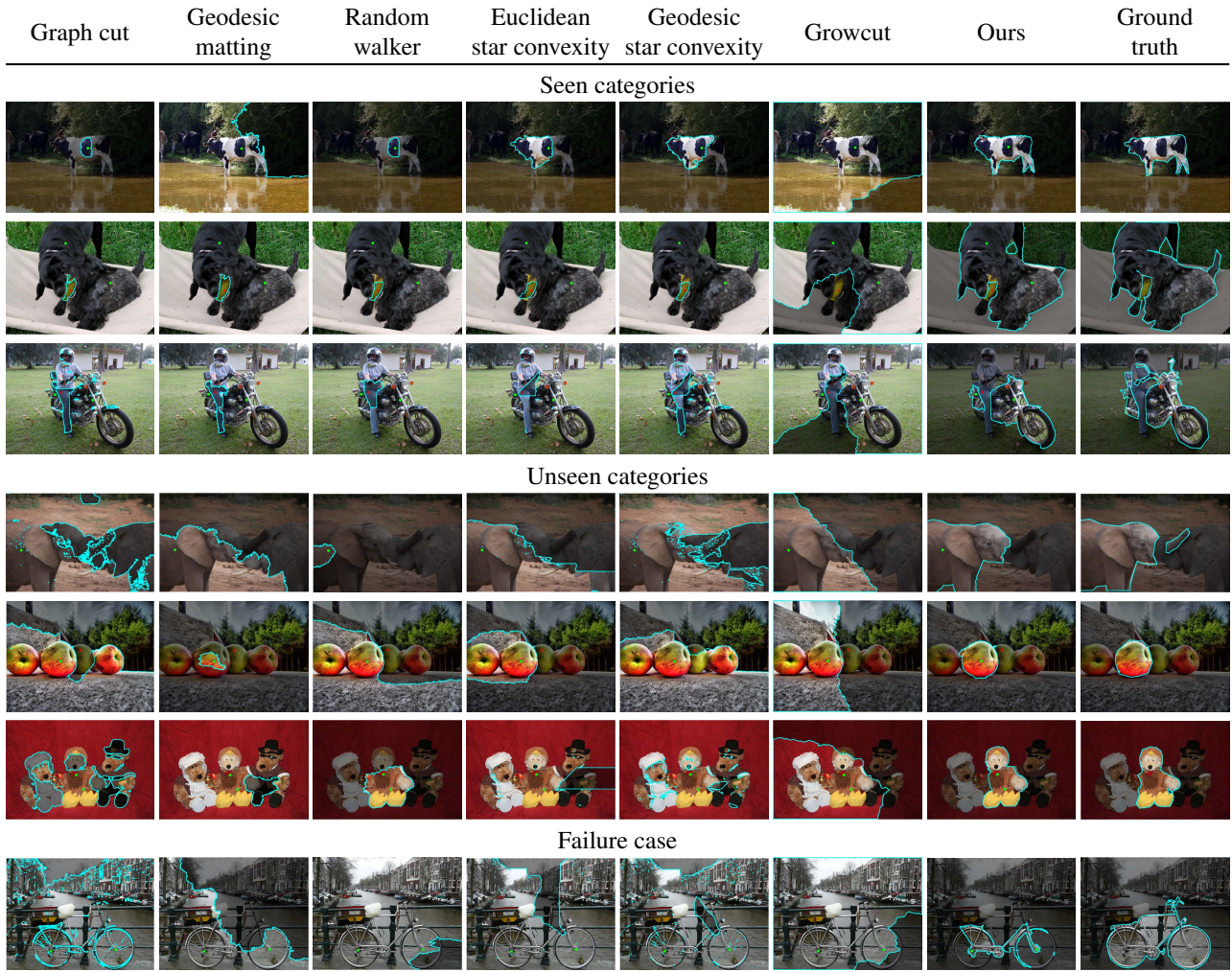

Figure 9: The segmentation results by different algorithms given the same user interaction sequences on the MS COCO dataset. The first to third rows are testing images from seen categories (*i.e.* "cow", "dog", "motorcycle"). The forth to sixth rows are testing images from unseen categories (*i.e.* "elephant", "apple", "teddy bear"). The last row is a failure example (*i.e.* "bicycle") by our algorithm. Each of the first seven columns represent the segmentation results by one algorithm and the rightmost column shows the ground truths. In each figure, green dots indicate positive clicks and red crosses indicate negative clicks. Background regions are faded to black and object boundaries are outlined in cyan.